\documentclass[10pt,twocolumn,letterpaper]{article}

\usepackage{cvpr}
\usepackage{times}
\usepackage{epsfig}
\usepackage{graphicx}
\usepackage{amsmath}
\usepackage{amssymb}

\usepackage{siunitx}
\usepackage{censor}

\usepackage[pagebackref=true,breaklinks=true,letterpaper=true,colorlinks,bookmarks=false]{hyperref}

\cvprfinalcopy 

\def\cvprPaperID{5} 

\ifcvprfinal\fi
\begin{document}

\title{Fine-Grained Head Pose Estimation Without Keypoints}

\author{Nataniel Ruiz~~~~~~Eunji Chong~~~~~~James M. Rehg\\
Georgia Institute of Technology\\
{\tt\small \{nataniel.ruiz, eunjichong, rehg\}@gatech.edu}
}

\maketitle

\begin{abstract}

Estimating the head pose of a person is a crucial problem that has a large amount of applications such as aiding in gaze estimation, modeling attention, fitting 3D models to video and performing face alignment. Traditionally head pose is computed by estimating some keypoints from the target face and solving the 2D to 3D correspondence problem with a mean human head model. We argue that this is a fragile method because it relies entirely on landmark detection performance, the extraneous head model and an ad-hoc fitting step. We present an elegant and robust way to determine pose by training a multi-loss convolutional neural network on 300W-LP, a large synthetically expanded dataset, to predict intrinsic Euler angles (yaw, pitch and roll) directly from image intensities through joint binned pose classification and regression. We present empirical tests on common in-the-wild pose benchmark datasets which show state-of-the-art results. Additionally we test our method on a dataset usually used for pose estimation using depth and start to close the gap with state-of-the-art depth pose methods. We open-source our training and testing code as well as release our pre-trained models~\footnote{\url{https://github.com/natanielruiz/deep-head-pose}}.

\end{abstract}

\section{INTRODUCTION}\label{sec1}
The related problems of head pose estimation and facial expression tracking have played an important role over the past 25 years in driving vision technologies for nonrigid registration and 3D reconstruction and enabling new ways to manipulate multimedia content and interact with users. Historically, there have been several major approaches to face modeling, with two primary ones being discriminative/landmark-based approaches~\cite{saragih2011deformable,Zhu2012} and parameterized appearance models, or PAMs~\cite{Cootes2001,Matthews2004} (see~\cite{Xiong2013} for additional discussion). In recent years, methods which directly extract 2D facial keypoints using modern deep learning tools~\cite{bulat2017far,zhu2016face,KEPLER} have become the dominant approach to facial expression analysis, due to their flexibility and robustness to occlusions and extreme pose changes. A by-product of keypoint-based facial expression analysis is the ability to recover the 3D pose of the head, by establishing correspondence between the keypoints and a 3D head model and performing alignment. However, in some applications the head pose may be all that needs to be estimated. In that case, is the keypoint-based approach still the best way forward? This question has not been thoroughly-addressed using modern deep learning tools, a gap in the literature that this paper attempts to fill.

We demonstrate that a direct, holistic approach to estimating 3D head pose from image intensities using convolutional neural networks delivers superior accuracy in comparison to keypoint-based methods. While keypoint detectors have recently improved dramatically due to deep learning, head pose recovery inherently is a two step process with numerous opportunities for error. First, if sufficient keypoints fail to be detected, then pose recovery is impossible. Second, the accuracy of the pose estimate depends upon the quality of the 3D head model. Generic head models can introduce errors for any given participant, and the process of deforming the head model to adapt to each participant requires significant amounts of data and can be computationally expensive.

While it is common for deep learning based methods using keypoints to jointly predict head pose along with facial landmarks, the goal in this case is to improve the accuracy of the facial landmark predictions, and the head pose branch is not sufficiently accurate on its own: for example \cite{KEPLER, ranjan2016hyperface, allinone} which are studied in Section \ref{sec4-A} and \ref{sec4-E}. A conv-net architecture which directly predicts head pose has the potential to be much simpler, more accurate, and faster. While other works have addressed the direct regression of pose from images using conv-nets~\cite{BMVC2015_130,patacchiola2017head,chang2017faceposenet} they did not include a comprehensive set of benchmarks or leverage modern deep architectures.

In applications where accurate head pose estimation is required, a common solution is to utilize RGBD (depth) cameras. These can be very accurate, but suffer from a number of limitations: First, because they use active sensing, they can be difficult to use outdoors and in uncontrolled environments, as the active illumination can be swamped by sunlight or ambient light. Second, depth cameras draw more power than RGB, resulting is significant battery life issues in mobile applications, and they are much less prevalent in general. Third, the data rates for RGBD are higher than for RGB, increasing storage and data transfer times. As a consequence, for a wide range of applications in domains such as pedestrian tracking and safety monitoring in autonomous driving, computer graphics, driver alertness monitoring, and social scene understanding from video, there remains a need for an RGB-based 3D head pose estimation solution which is fast and reliable.

The key contributions of our work are the following:
\begin{itemize}
\itemsep0em 
\item Proposing a method to predict head pose Euler angles directly from image intensities using a multi loss network which has a loss for each angle and each loss has two components: a pose bin classification and a regression component. We outperform published methods in single frame pose estimation in several datasets.
\item Demonstrating the generalization capacity of our model by training it on a large synthetic dataset and obtaining good results on several testing datasets.
\item Presenting ablation studies on the convolutional architecture of the network as well as on the multiple components of our loss function.
\item Presenting a detailed study of the accuracy of pose from 2D landmark methods, and detail weaknesses of this approach which are solved by the appearance based approach that we take.
\item Studying the effects of low resolution on pose estimation for different methods. We show that our method coupled with data augmentation is effective in tackling the interesting problem of head pose estimation on low resolution images.
\end{itemize}

\section{RELATED WORK}\label{sec2}

Human head pose estimation is a widely studied task in computer vision with very diverse approaches throughout its history. In the classic literature we can discern Appearance Template Models which seek to compare test images with a set of pose exemplars~\cite{s88, s110, s111}. Detector arrays were once a popular method when frontal face detection~\cite{s97,s104} had increased success, the idea was to train multiple face detectors for different head poses~\cite{s47, s146}.

Recently, facial landmark detectors which have become very accurate~\cite{bulat2017far,zhu2016face,KEPLER}, have been popular for the task of pose estimation.

Also recently, work has developed on estimating head pose using neural networks. \cite{patacchiola2017head} presents an in-depth study of relatively shallow networks trained using a regression loss on the AFLW dataset. In KEPLER~\cite{KEPLER} the authors present a modified GoogleNet architecture which predicts facial keypoints and pose jointly. They use the coarse pose supervision from the AFLW dataset in order to improve landmark detection. Two works dwell on building one network to fulfill various prediction tasks regarding facial analysis. Hyperface~\cite{ranjan2016hyperface} is a CNN that sets out to detect faces, determine gender, find landmarks and estimate head pose at once. It does this by using an R-CNN~\cite{girshick14CVPR} based approach and a modified AlexNet architecture which fuses intermediate convolutional layer outputs and adds separate fully-connected networks to predict each subtask. All-In-One Convolutional Neural Network~\cite{allinone} for Face Analysis adds smile, age estimation and facial recognition to the former prediction tasks. We compare our results to all of these works.

Chang et al.~\cite{chang2017faceposenet} also argue for landmark-free head pose estimation. They regress 3D head pose using a simple CNN and focus on facial alignment using the predicted head pose. They demonstrate the success of their approach by improving facial recognition accuracy using their facial alignment pipeline. They do not directly evaluate their head pose estimation results. This differs from our work since we directly evaluate and compare our head pose results extensively on annotated datasets.

Work from Gu et al.\cite{nvidia} uses a VGG network to regress the head pose Euler angles. Instead of improving single-frame prediction by modifying the network structure it focuses on using a recurrent neural network to improve pose prediction by leveraging the time dimension which we do not use. They evaluate their work on a synthetic dataset as well as a real-world dataset. Another key difference with our work is that we set out to show generalization capacity of our network by training on a large dataset and testing the performance of that network on various external datasets without finetuning the network on those datasets. We believe this is a good way to measure how the model will generalize in real applications.


\section{METHOD}\label{sec3}

In this section we describe the advantages of estimating head pose with deep networks directly from image intensities and argue that it should be preferred to landmark-to-pose methods. We explain how combined classification and regression can be used to improve performance when training on the larger synthetic 300W-LP~\cite{zhu2016face} dataset. We also talk about key insights regarding data augmentation, training and testing datasets and how to improve performance for low-resolution images.

\subsection{Advantages of Deep Learning for Head Pose Estimation}\label{sec3-A}

Even though it might seem evident to the reader that given careful training deep networks can accurately predict head pose this approach has not been studied extensively and is not commonly used for head pose estimation tasks. Instead if very accurate head pose is needed then depth cameras are installed and if no depth footage exists landmarks are detected and pose is retrieved. In this work we show that a network trained on a large synthetic dataset, which by definition has accurate pose annotations, can predict pose accurately in real cases. We test the networks on real datasets which have accurate pose annotations and show state-of-the-art results on the AFLW, AFLW2000~\cite{zhu2016face} and BIWI~\cite{fanelli_IJCV} datasets. Additionally we are starting to close the gap with very accurate methods which use depth information on the BIWI dataset.

We believe that deep networks have large advantages compared to landmark-to-pose methods, for example:
\begin{itemize}
\itemsep0em 
  \item They are not dependent on: the head model chosen, the landmark detection method, the subset of points used for alignment of the head model or the optimization method used for aligning 2D to 3D points.
  
  \item They always output a pose prediction which is not the case for the latter method when the landmark detection method fails. 
\end{itemize}

\begin{figure}[t]
\begin{center}
   \includegraphics[width=1\linewidth]{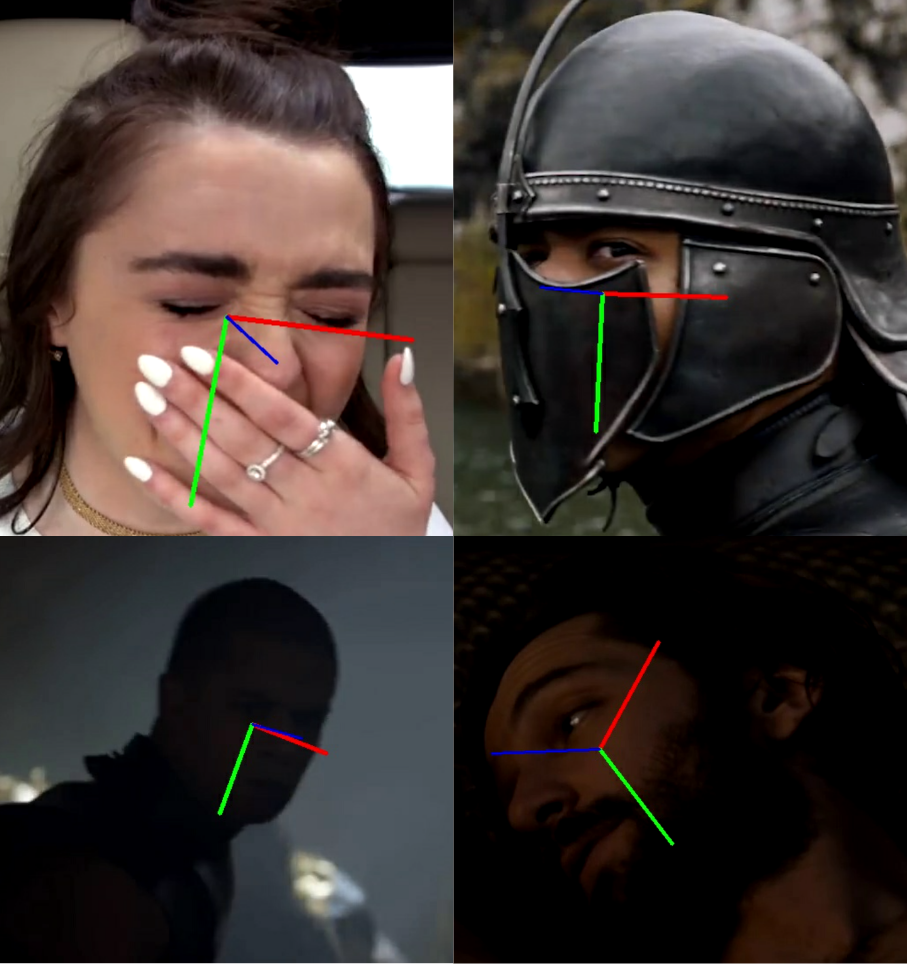}
\end{center}
   \caption{Example pose detections in difficult scenarios using our proposed method. The blue axis points towards the front of the face, green pointing downward and red pointing to the side. Best viewed in color.}
   \label{examples_1}
\end{figure}
\

\subsection{The Multi-Loss Approach}\label{sec3-B}

All previous work which predicted head pose using convolutional networks regressed all three Euler angles directly using a mean squared error loss. We notice that this approach does not achieve the best results on our large-scale synthetic training data.

We propose to use three separate losses, one for each angle. Each loss is a combination of two components: a binned pose classification and a regression component. Any backbone network can be used and augmented with three fully-connected layers which predict the angles. These three fully-connected layers share the previous convolutional layers of the network.

\begin{figure*}
  \includegraphics[width=\textwidth,height=6.5cm]{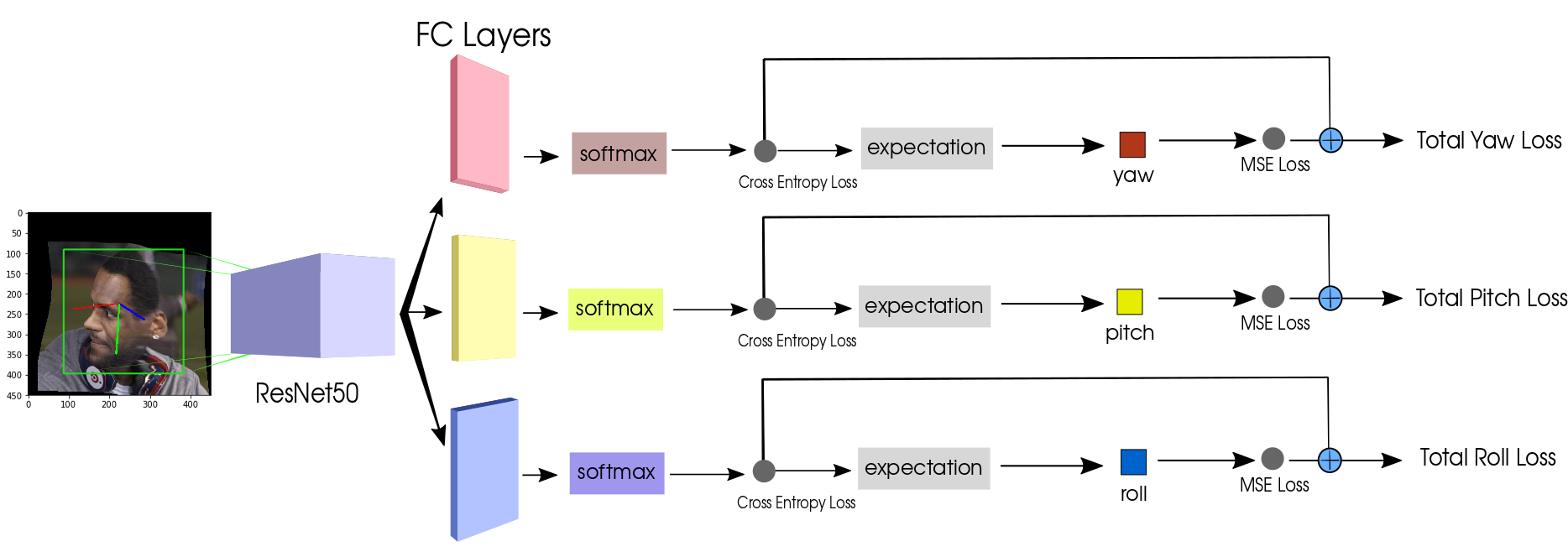}
  \caption{ResNet50 architecture with combined Mean Squared Error and Cross Entropy Losses.}
  \label{architecture}
\end{figure*}

The idea behind this approach is that by performing bin classification we use the very stable softmax layer and cross-entropy, thus the network learns to predict the neighbourhood of the pose in a robust fashion. By having three cross-entropy losses, one for each Euler angle, we have three signals which are backpropagated into the network which improves learning. In order to obtain a fine-grained predictions we compute the expectation of each output angle for the binned output. The detailed architecture is shown in Figure \ref{architecture}.

We then add a regression loss to the network, namely a mean-squared error loss, in order to improve fine-grained predictions. We have three final losses, one for each angle, and each is a linear combination of both the respective classification and the regression losses. We vary the weight of the regression loss in Section \ref{sec4-F} and we hold the weight of the classification loss constant at $1$. The final loss for each Euler angle is the following:

\[\boldsymbol{\mathcal{L}}=H(y, \hat{y}) + \alpha \cdot MSE(y,\hat{y}) \]

Where $H$ and $MSE$ respectively designate the cross-entropy and mean squared error loss functions.\\

We experiment with different coefficients for the regression loss and present our results in Section \ref{sec4-F}.

\subsection{Datasets for Fine-Grained Pose Estimation}\label{sec3-C}

In order to truly make progress in the problem of predicting pose from image intensities we have to find real datasets which contain precise pose annotations, numerous identities, different lighting conditions, all of this across large poses. We identify two very different datasets which fill these requirements.

First is the challenging AFLW2000 dataset. This dataset contains the first 2000 identities of the in-the-wild AFLW dataset which have been re-annotated with 68 3D landmarks using a 3D model which is fit to each face. Consequently this dataset contains accurate fine-grained pose annotations and is a prime candidate to be used as a test set for our task.

Second the BIWI dataset is gathered in a laboratory setting by recording RGB-D video of different subjects across different head poses using a Kinect v2 device. It contains roughly 15,000 frames and the rotations are $\pm 75^\circ$ for yaw, $\pm 60^\circ$ for pitch and $\pm 50^\circ$ for roll. A 3D model was fit to each individual's point cloud and the head rotations were tracked to produce the pose annotations. This dataset is commonly used as a benchmark for pose estimation using depth methods which attests to the precision of its labels. In our case we will not use the depth information nor the temporal information, only individual color frames. In Section \ref{sec4-A} we compare to a very accurate state-of-the-art depth method to ascertain the performance gap between approaches.

\subsection{Training on a Synthetically Expanded Dataset}\label{sec3-D}

We follow the path of \cite{bulat2017far} which used synthetically expanded data to train their landmark detection model. One of the datasets they train on is the 300W-LP dataset which is a collection of popular in-the-wild 2D landmark datasets which have been grouped and re-annotated. A face model is fit on each image and the image is distorted to vary the yaw of the face which gives us pose across several yaw angles. Pose is accurately labeled because we have the 3D model and 6-D degrees of freedom of the face for each image. 

We show in Section \ref{sec4-A} that by carefully training on large amounts of synthetic data we can begin closing the gap with existing depth methods and can achieve very good accuracies on datasets with fine-grained pose annotations. We also test our method against other deep learning methods whose authors have graciously run on some of the test datasets that we use in Section \ref{sec4-A}. Additionally in the same Section, we test landmark-to-pose methods and other types of pose estimation methods such as 3D model fitting.

\subsection{The Effects of Low-Resolution}\label{sec3-E}
Currently there is need for head pose estimation at a distance and there exist multiple example applications in areas such as video surveillance, autonomous driving and advertisement. Future head pose estimation methods should look to improve estimation for low-resolution heads.

We present an in-depth study of the effect of low-resolution on widely-used landmark detectors as well as state-of-the-art detectors. We contend that low-resolution should worsen the performance of landmark detection since estimating keypoints necessitates access to features which disappear at lower resolutions. We argue that although detailed features are important for pose estimation they are not as critical. Moreover this area is relatively untapped: there is scarce related work discussing head pose estimation at a distance. As far as we know there is no work discussing low-resolution head pose estimation using deep learning.

Deep networks which predict pose directly from image intensities are a good candidate method for this application because robustness can be built into them by modifying the network or augmenting its training data in smart ways. We propose a simple yet surprisingly effective way of developing robustness to low-resolution images: we augment our data by downsampling and upsampling randomly which forces the network to learn effective representations for varied resolutions. We also augment the data by blurring the images. Experiments are shown in Section \ref{sec4-F}

\section{EXPERIMENTAL RESULTS}\label{sec4}
We perform experiments showing the overall performance of our proposed method on different datasets for pose estimation as well as popular landmark detection datasets. We show ablation studies for the multi-loss. Additionally, we delve into landmark-to-pose methods and shed light on their robustness. Finally we present experiments suggesting that a holistic approach to pose using deep networks outperforms landmark-to-pose methods when resolution is low even if the landmark detector is state-of-the-art.

\subsection{Fine-Grained Pose Estimation on the AFLW2000 and BIWI Datasets}\label{sec4-A}
We evaluate our method on the AFLW2000 and BIWI datasets for the task of fine-grained pose estimation and compare to pose estimated from landmarks using two different landmark detectors, FAN~\cite{bulat2017far} and Dlib~\cite{kazemi2014one}, and ground-truth landmarks (only available for AFLW2000).

FAN is a very impressive state-of-the-art landmark detector described in~\cite{bulat2017far} by Bulat and Tzimiropoulos. It uses Stacked Hourglass Networks~\cite{newell2016stacked} originally intended for human body pose estimation and switches the normal ResNet Bottleneck Block for a hierarchical, parallel and multi-scale block proposed in another paper by the same authors~\cite{bulat2017binarized}. We were inspired to train our pose-estimation network on 300W-LP from their work which trains their network on this dataset for the task of landmark detection.
Dlib implements a landmark detector which uses an ensemble of regression trees and which is described in~\cite{kazemi2014one}.

We run both of these landmark detectors on the AFLW2000 and BIWI datasets. AFLW2000 images are small and are cropped around the face. For BIWI we run a Faster R-CNN~\cite{renNIPS15fasterrcnn} face detector trained on the WIDER Face Dataset~\cite{yang2016wider,jiang2017face} and deployed in a Docker container~\cite{ruiz2017dockerface}. We loosely crop the faces around the bounding box in order to conserve the rest of the head. We also retrieve pose from the ground-truth landmarks of AFLW2000. Results can be seen in Tables \ref{raw_performance_AFLW2000} and \ref{raw_performance_BIWI}.

Additionally, we run 3DDFA~\cite{zhu2016face} which directly fits a 3D face model to RGB image via convolutional neutral networks. The primary task of 3DDFA is to align facial landmarks even for the occluded ones using a dense 3D model. As a result of their 3D fitting process, a 3D head pose is produced and we report this pose.

Finally, we compare our results to the state-of-the-art RGBD method~\cite{3DMM}. We can see that our proposed method considerably shrinks the gap between RGBD methods and ResNet50~\cite{He2015}. Pitch estimation is still lagging behind in part due to the lack of large quantities of extreme pitch examples in the 300W-LP dataset. We expect that this gap will be closed when more data is available.

We present two multi-loss ResNet50 networks with different regression coefficients of $1$ and $2$ trained on the 300W-LP dataset. For BIWI we also present a multi-loss ResNet50 ($\alpha=1$) trained on AFLW. All three networks were trained for 25 epochs using Adam optimization\cite{kingma2014adam} with a learning rate of $10^{-5}$ and $\beta_{1}=0.9, \beta_{2}=0.999$ and $\epsilon=10^{-8}$. We normalize the data before training by using the ImageNet mean and standard deviation for each color channel. Note that since our method bins angles in the $\pm 99^\circ$ range we discard images with angles outside of this range. Only 31 images are not used from the 2000 images of AFLW2000.

In order to compare to Gu et al.~\cite{nvidia} we train on three different 70-30 splits of videos in the BIWI dataset and we average our mean average error for each split. For this evaluation we use weight decay with a coefficient of $0.04$ because of the smaller amount of data available. We compare our result to their single-frame result which was trained in the same fashion and we show the results in Table \ref{NVIDIA_comparison}. Our method compares favorably to Gu et al. and lowers the sum of mean average errors by $1.29^\circ$.

\begin{table}[]
\centering
\resizebox{\columnwidth}{!}{%

\begin{tabular}{lllll}
                               & Yaw    & Pitch  & Roll                         & MAE    \\
Multi-Loss ResNet50 ($\alpha=2$)              & 6.470  & 6.559  & 5.436                        & \bf{6.155}  \\
Multi-Loss ResNet50 ($\alpha=1$)              & 6.920  & 6.637  & 5.674                        & 6.410  \\
3DDFA~\cite{zhu2016face}  & 5.400  & 8.530  & 8.250                        & 7.393  \\
FAN~\cite{bulat2017far} (12 points)                & 6.358  & 12.277 & 8.714                        & 9.116  \\
Dlib~\cite{kazemi2014one} (68 points)               & 23.153 & 13.633 & 10.545                       & 15.777 \\
Ground truth landmarks         & 5.924  & 11.756 & 8.271                        & 8.651
\end{tabular}
}
\caption{Mean average error of Euler angles across different methods on the AFLW2000 dataset~\cite{zhu2016face}.}
\label{raw_performance_AFLW2000}
\end{table}

\begin{table}[]
\centering
\resizebox{\columnwidth}{!}{%
\begin{tabular}{lllll}
                        & Yaw    & Pitch  & Roll                         & MAE    \\
Multi-Loss ResNet50 ($\alpha=2$)       & 5.167  & 6.975  & 3.388                        & 5.177  \\
Multi-Loss ResNet50 ($\alpha=1$)       & 4.810  & 6.606  & 3.269                        & \bf{4.895}  \\
KEPLER~\cite{KEPLER}$\dagger$                  & 8.084  & 17.277 & 16.196                       & 13.852 \\
Multi-Loss ResNet50 ($\alpha=1$)$\dagger$ & 5.785  & 11.726 & 8.194                        & 8.568  \\
3DMM+ Online~\cite{3DMM} \text{*}     & 2.500  & 1.500  & 2.200                        & \bf{2.066}  \\
FAN~\cite{bulat2017far} (12 points)         & 8.532  & 7.483  & 7.631                        & 7.882  \\
Dlib~\cite{kazemi2014one} (68 points)        & 16.756 & 13.802 & 6.190                        & 12.249 \\
3DDFA~\cite{zhu2016face}                   & 36.175 & 12.252 & 8.776                        & 19.068
\end{tabular}
}
\caption{Mean average error of Euler angles across different methods on the BIWI dataset~\cite{fanelli_IJCV}. \text{*} These methods use depth information. $\dagger$ Trained on AFLW}
\label{raw_performance_BIWI}
\end{table}

\begin{table}[]
\centering
\resizebox{\columnwidth}{!}{%
\begin{tabular}{lllll}
                               & Yaw    & Pitch  & Roll                         & Sum of errors    \\
Multi-Loss ResNet50 ($\alpha=1$)              & \bf{3.29}  & \bf{3.39}  & \bf{3.00}                        & \bf{9.68}  \\
Gu et al.~\cite{nvidia}  & 3.91  & 4.03  & 3.03                        & 10.97  \\
\end{tabular}
}
\caption{Comparison with Gu et al.~\cite{nvidia}. Mean average error of Euler angles averaged over train-test splits of the BIWI dataset~\cite{fanelli_IJCV}.}
\label{NVIDIA_comparison}
\end{table}

\subsection{Landmark-To-Pose Study}\label{sec4-B}
In this set of experiments, we examine the approach of using facial landmarks as a proxy to head pose and investigate the limitations of its use for pose estimation. The commonly used pipeline for landmark-to-pose estimation involves a number of steps; 2D landmarks are detected, 3D human mean face model is assumed, camera intrinsic parameters are approximated, and finally the 2D-3D correspondence problem is solved. We show how this pipeline is affected by different error sources. Specifically, using the AFLW2000 benchmark dataset, we conduct experiments starting from the best available condition (ground truth 2D landmarks, ground truth 3D mean face model) and examine the final head pose estimation error by deviating from this condition. For all of these experiments, we assume zero lens distortion, and run iterative method based on Levenberg-Marquardt optimization to find 2D-3D correspondence which is implemented as the function SolvePnP in OpenCV.

\begin{figure}[t]
\begin{center}
   \includegraphics[width=1\linewidth]{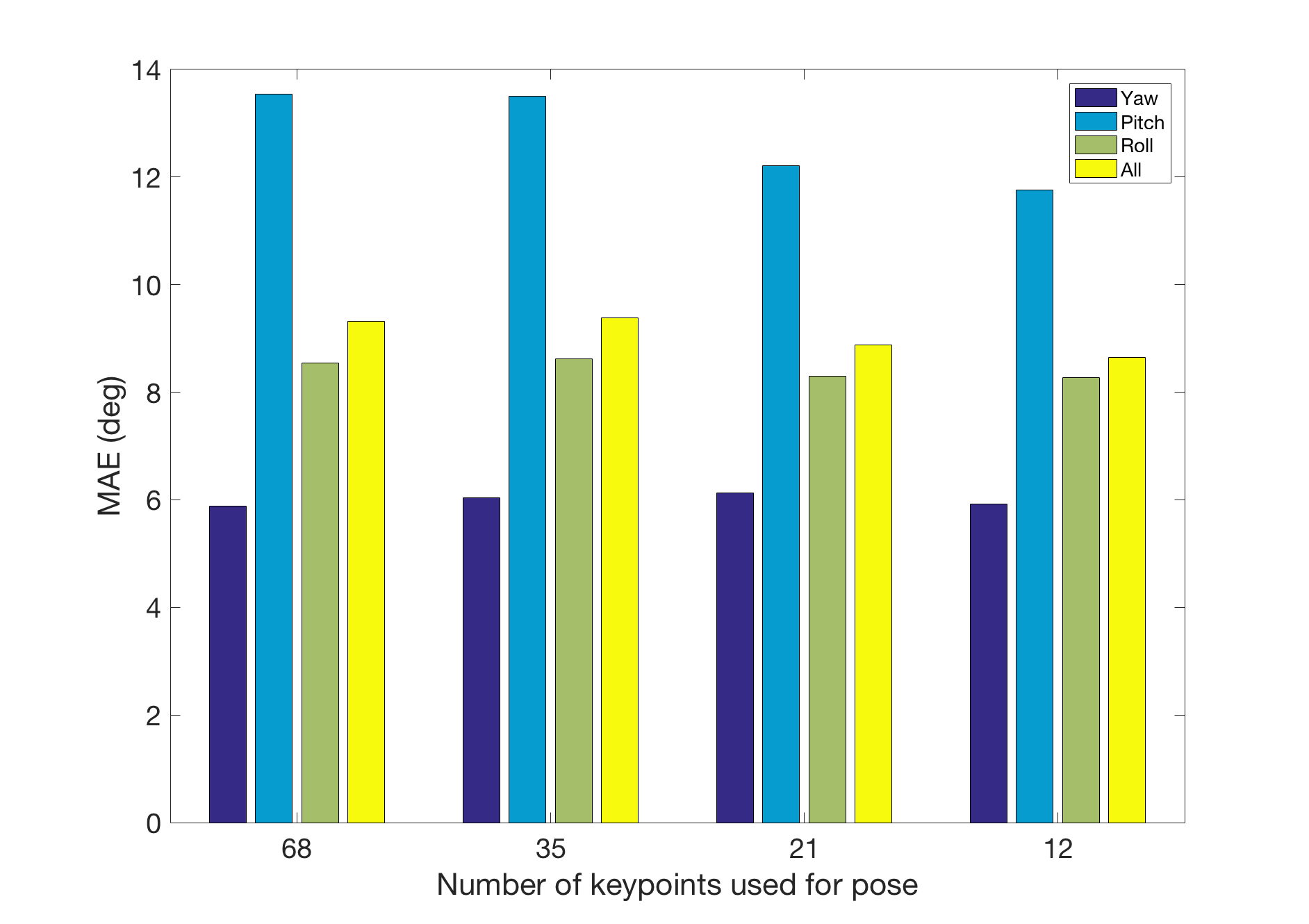}
\end{center}
   \caption{We show the effects of using different number of landmark points for 3D head pose estimation using ground truth facial landmarks and the ground truth mean face model on the AFLW2000 dataset.}
   \label{num_keypoints}
\end{figure}

\begin{figure*}[t]
\begin{center}
   \includegraphics[width=0.24\linewidth]{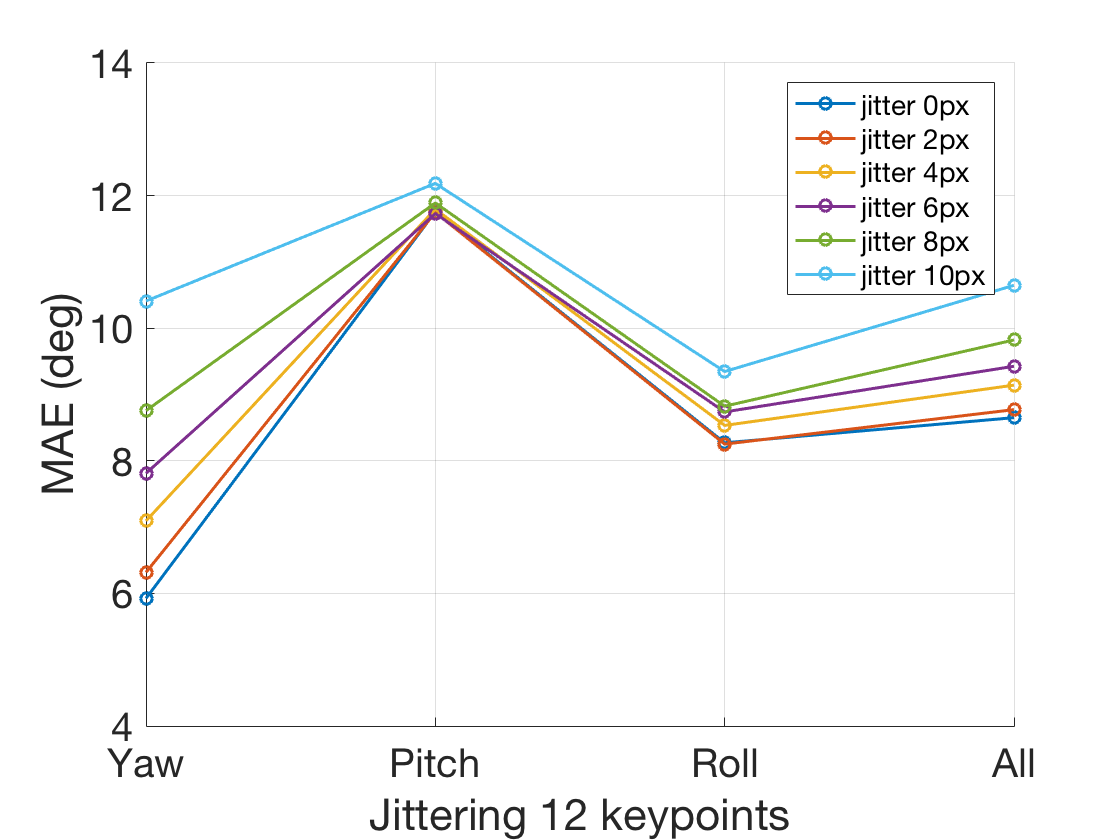}
   \includegraphics[width=0.24\linewidth]{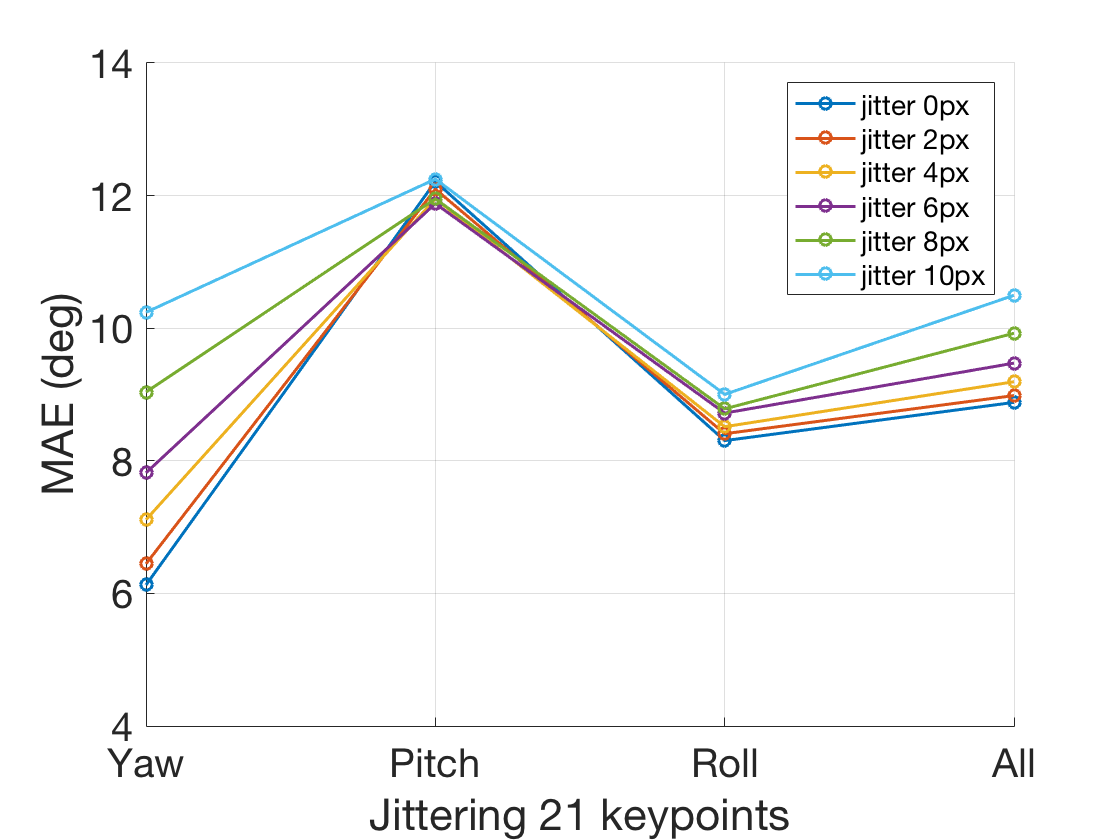}
   \includegraphics[width=0.24\linewidth]{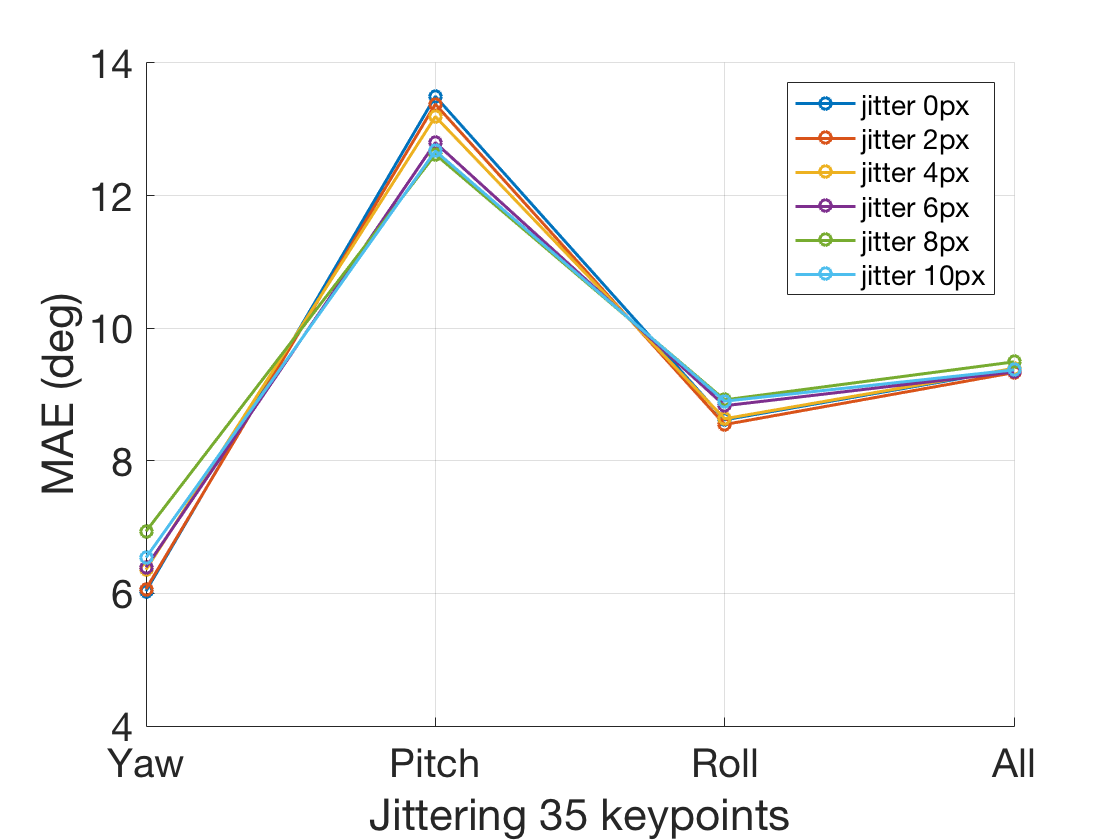}
   \includegraphics[width=0.24\linewidth]{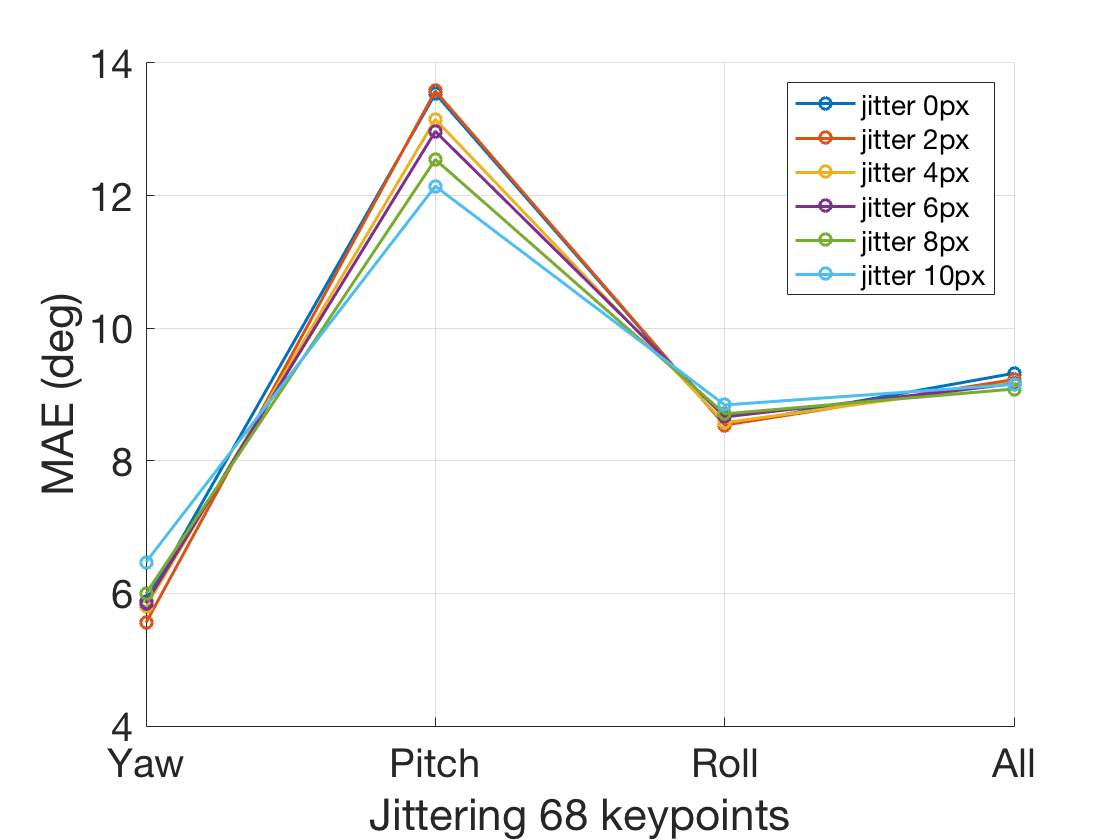}
\end{center}
   \caption{We show the effect of jittering landmark points around their ground truth position on the task of 3D head pose estimation on AFLW2000 to simulate the effects of noise in the facial keypoint detector. We repeat this experiment four times with different number of landmarks. For all experiments we use the ground truth mean face model for the landmark-to-pose alignment task.}
   \label{jittering}
\end{figure*}

\begin{figure*}[t]
\begin{center}
   \includegraphics[width=0.4\linewidth]{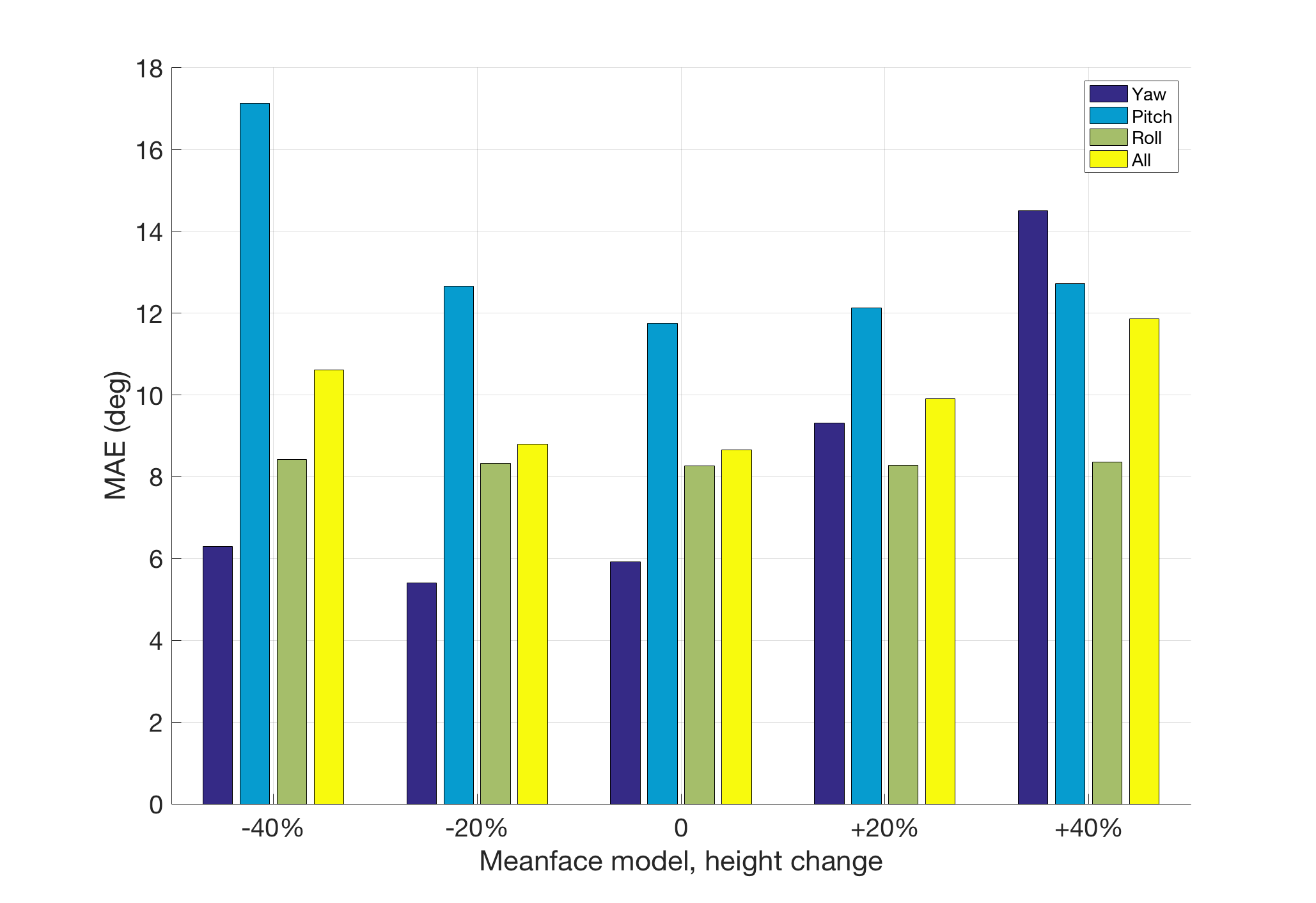}
   \includegraphics[width=0.4\linewidth]{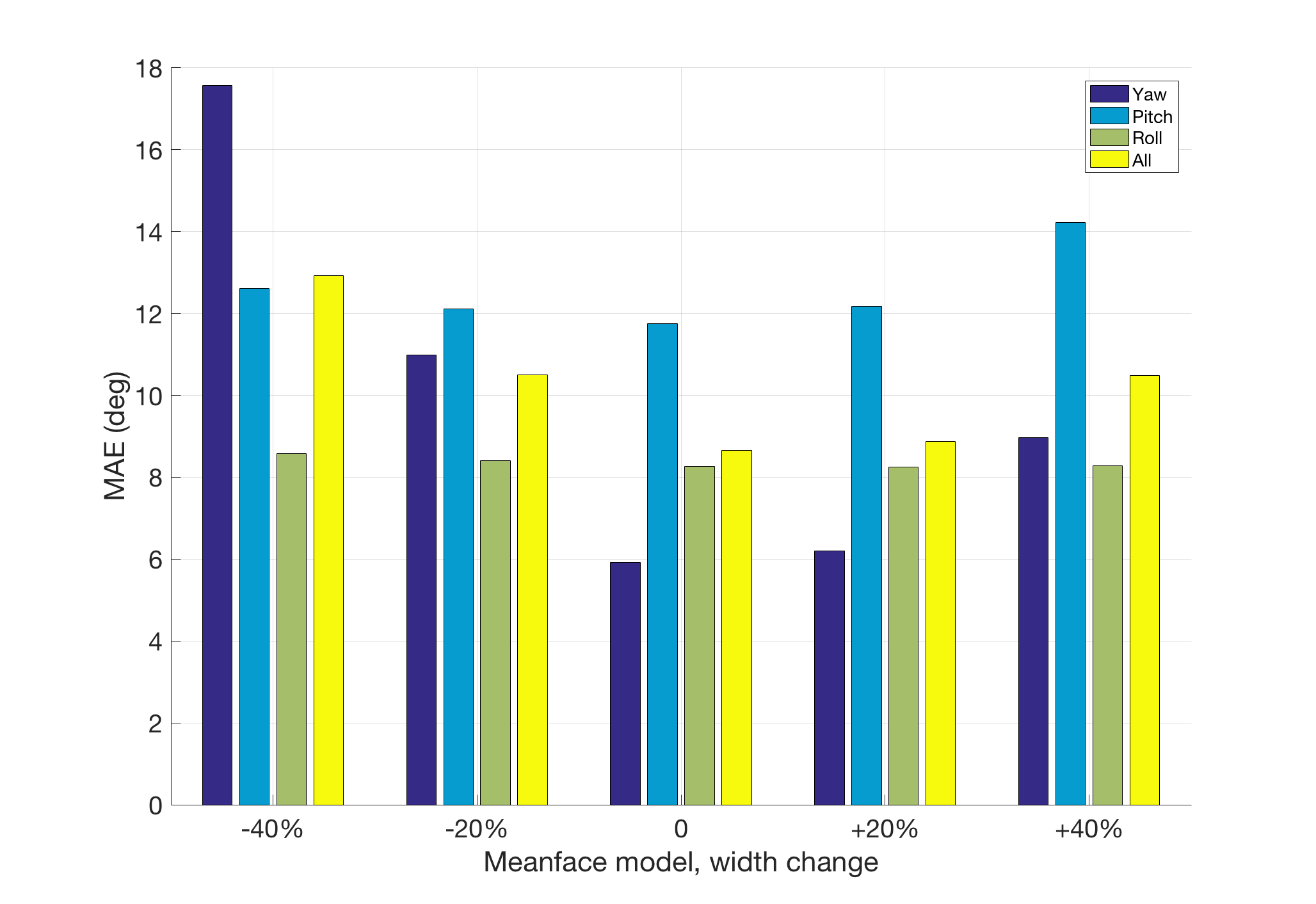}
\end{center}
   \caption{We show the effects of changing the 3D mean face model on the task of 3D head pose estimation from 2D landmarks. We use 2D ground truth landmarks and modify the mean face model by stretching its width and height.}
   \label{meanface}
\end{figure*}

\begin{figure*}[t]
\begin{center}
   \includegraphics[width=0.4\linewidth]{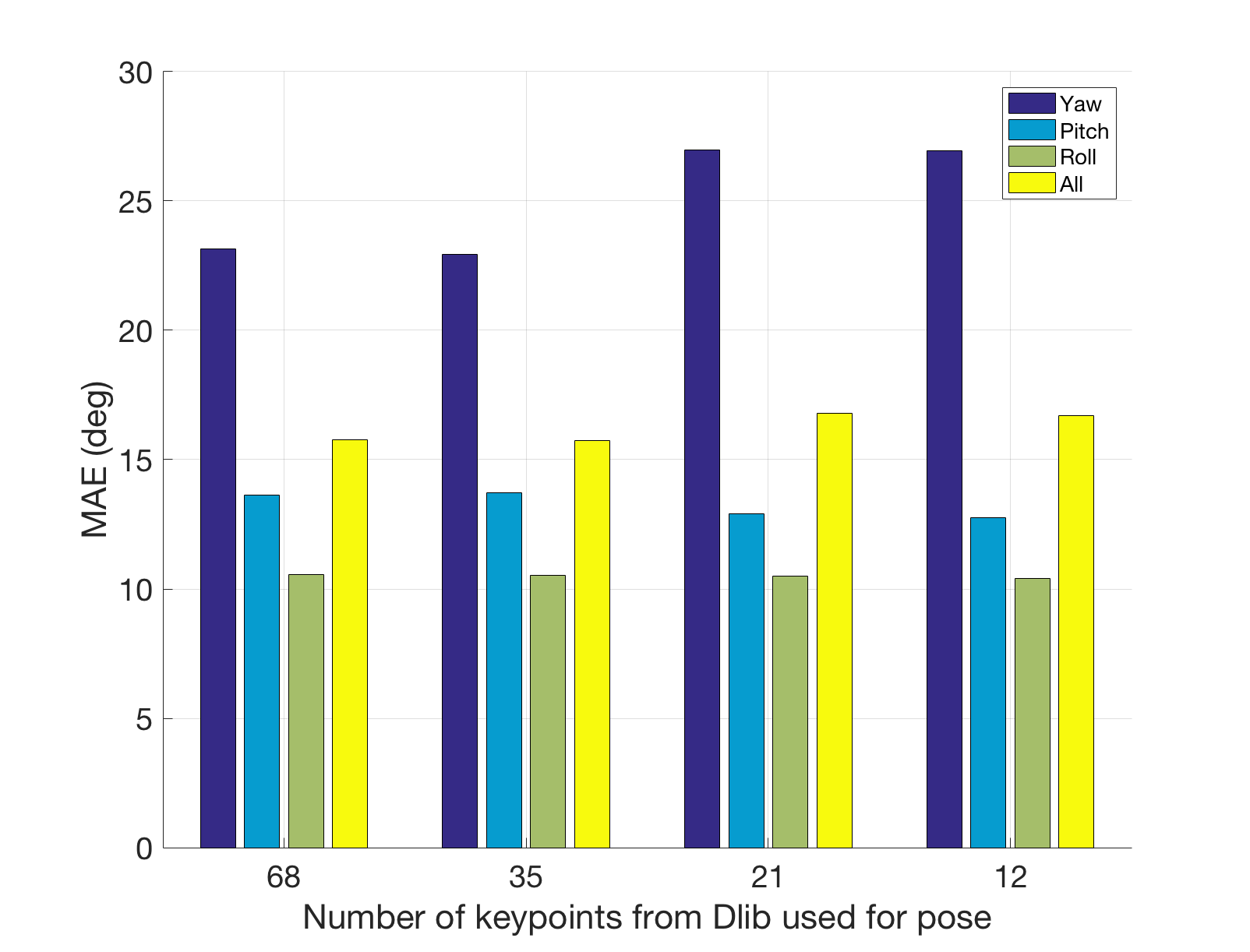}
   \includegraphics[width=0.4\linewidth]{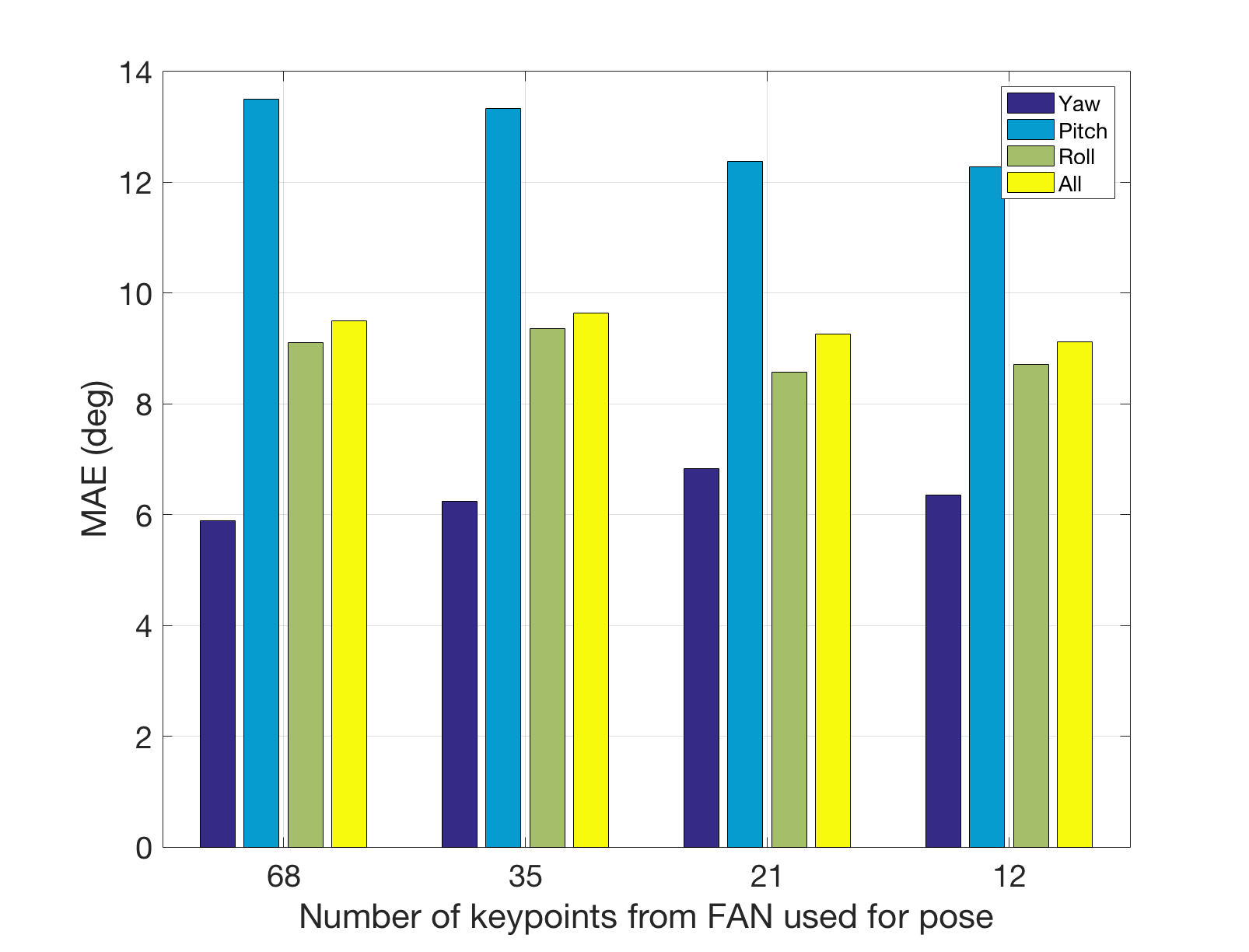}
\end{center}
   \caption{Using estimated 2D landmark points, this experiment shows the 3D pose estimation error depending on how many facial keypoints are used.}
   \label{estimated_2to3}
\end{figure*}

We first run the pipeline only with ground truth landmarks, varying the number of points used in the optimization method. We observe that in this ideal condition, using all of the available 68 landmark points actually gives biggest error as shown in Figure~\ref{num_keypoints}. Then, we jitter the ground truth 2D landmarks by adding random noise independently in x, y direction per landmark. Figure~\ref{jittering} shows the results of this experiment with up to 10 pixel of jittering. We repeat the experiment with the same set of keypoints selected for Figure~\ref{num_keypoints}. Finally, we change the mean face model by stretching the ground truth mean face in width and height up to 40\% Figure~\ref{meanface}. Additionally, we also report results based on estimated landmarks using FAN and Dlib in Figure~\ref{estimated_2to3}.

The results suggest that with ground truth 2D landmarks, using less key points produces less error since it's less likely to be affected by pose-irrelevant deformation such as facial expression. However, the more points we use for correspondence problem, the more robust it becomes to random jittering. In other words, there exists a tradeoff; if we know the keypoints are very accurate we want to use less points for pose, but if there's error we want to use more points. With \textit{estimated} landmarks, it's not clear how we can weigh these two, and we find that using more points can both help and worsen pose estimation as presented in Figure~\ref{estimated_2to3}.

\subsection{AFLW and AFW Benchmarking}\label{sec4-E}
\begin{figure}[t]
\begin{center}
   \includegraphics[width=1\linewidth]{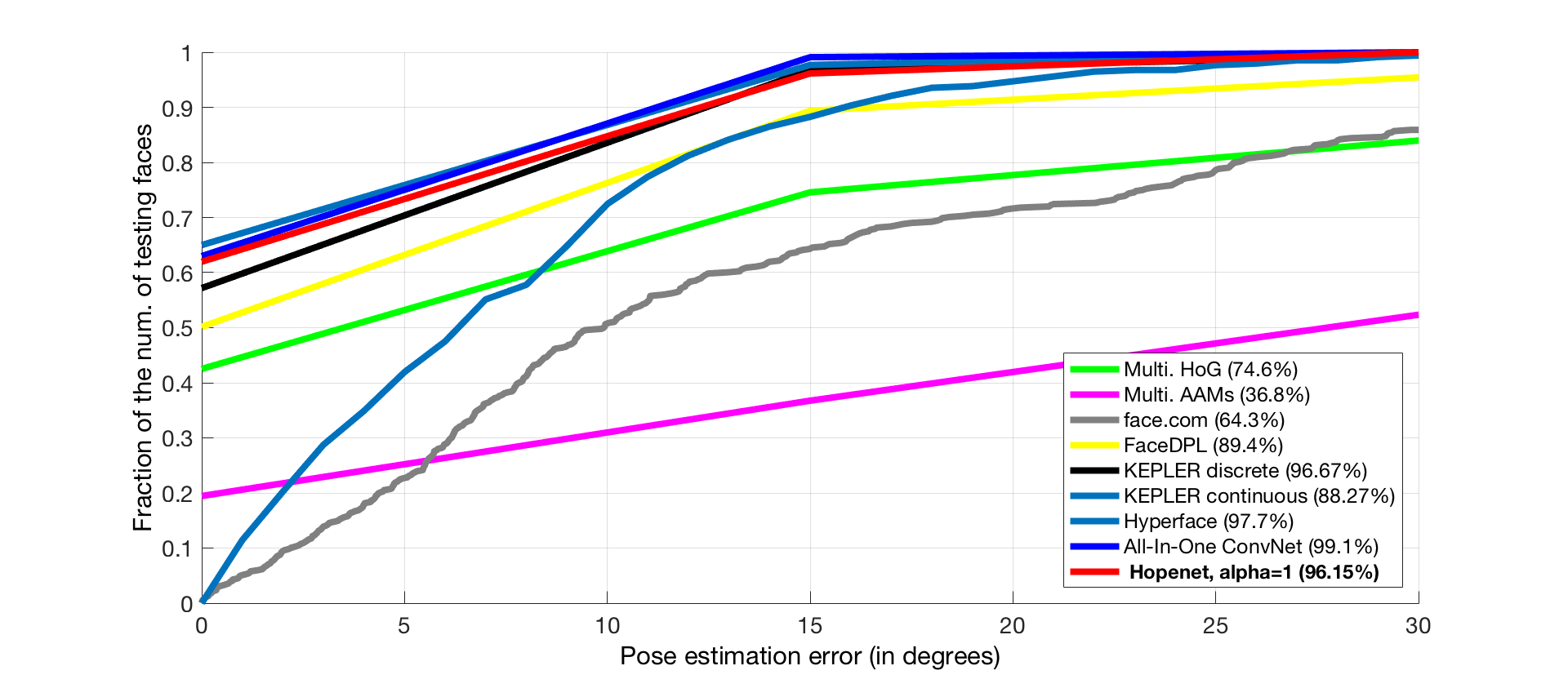}
\end{center}
   \caption{AFW pose benchmark result along with other methods~\cite{allinone,ranjan2016hyperface,KEPLER,zhu2012face}.}
   \label{afw_pose}
\end{figure}
The AFLW dataset, which is commonly used to train and test landmark detection methods, also includes pose annotations. Pose was obtained by annotating landmarks and using a landmark-to-pose method. Results can be seen in Table \ref{AFLW}.

AFW is a popular dataset, also commonly used to test landmark detection, which contains rough pose annotations. It contains 468 in-the-wild faces with absolute yaw degree's up to $\pm 90^\circ$. Methods only compare mean average error for yaw. Methods usually output discrete predictions and round their output to the closest $15^\circ$ multiple. As such at the $15^\circ$ error margin, which is one of the main metrics reported in the literature, this dataset is saturated and methods achieve over 95\% accuracy. Results are shown in Figure \ref{afw_pose}.

Using our joint classification and regression losses for AlexNet~\cite{krizhevsky2012imagenet} we obtain similar mean average error after training for 25 epochs. We compare our results to the KEPLER~\cite{KEPLER} method which uses a modified GoogleNet for simultaneous landmark detection and pose estimation and to \cite{patacchiola2017head} which uses a 4-layer convolutional network. Multi-Loss ResNet50 achieves lower Mean Average Error than KEPLER across all angles in the AFLW test-set after 25 epochs of training using Adam and same learning parameters as in Section \ref{sec4-A}. These results can be observed in Table \ref{AFLW}.

We test the previously trained AlexNet and Multi-Loss ResNet50 networks on the AFW dataset and display the results in Figure~\ref{afw_pose}. We evaluate the results uniquely on the yaw as all related work does. We constrain our networks to output discrete yaw in 15 degree increments and display the accuracy at two different yaw thresholds. A face is correctly classified if the absolute error of the predicted yaw is lower or equal than the threshold presented.

The same testing protocol is adopted for all compared methods and numbers are reported directly from the associated papers. Hyperface~\cite{ranjan2016hyperface} and All-In-One~\cite{allinone} both use a single network for numerous facial analysis tasks. Hyperface uses an AlexNet pre-trained on ImageNet as a backbone and All-In-One uses a backbone 7-layer conv-net pre-trained on the face recognition task using triplet probability constraints \cite{sankaranarayanan2016triplet}.

We show that by pre-training on ImageNet and fine-tuning on the AFLW dataset we achieve accuracies that are very close to the best results of the related work. We do not use any other supervisory information which might improve the performance of the network such as 2D landmark annotations. We do however use a more powerful backbone network in ResNet50. We show performance of the same network on both the AFLW test-set and AFW.

\begin{table}[]
\centering
\resizebox{\columnwidth}{!}{%
\begin{tabular}{lllll}
              & Yaw  & Pitch & Roll & MAE   \\
Multi-Loss ResNet50 ($\alpha=1$) & 6.26 & 5.89  & 3.82 & \bf{5.324} \\
AlexNet ($\alpha=1$) & 7.79 & 7.41  & 6.05 & 7.084 \\
KEPLER~\cite{KEPLER}        & 6.45 & 5.85  & 8.75 & 7.017 \\
Patacchiola, Cangelosi~\cite{patacchiola2017head} & 11.04 & 7.15  & 4.4 & 7.530 \\
\end{tabular}
}
\caption{Mean average errors of predicted Euler angles in the AFLW test set.}
\label{AFLW}
\end{table}

\subsection{AFLW2000 Multi-Loss Ablation}\label{sec4-F}
In this section we present an ablation study of the multi-loss. We train ResNet50 only using a Mean Squared Error (MSE) Loss and compare this to ResNet50 using a multi-loss with different coefficients for the MSE component. The weight of the Cross-Entropy loss is maintained constant at 1. We also compare this to AlexNet to discern the effects of having a more powerful architecture.

We observe the best results on the AFLW2000 dataset when the regression coefficient is equal to 2. We demonstrate increased accuracy when weighing each loss roughly with the same magnitude. This phenomenon can be observed in Table \ref{AFLW2000_ablation}.

\begin{table}[]
\centering
\resizebox{\columnwidth}{!}{%

\begin{tabular}{llllll}
                         & $\alpha$  & Yaw    & Pitch & Roll  & MAE    \\
ResNet50 regression only &             & 13.110 & 6.726 & 5.799 & 8.545  \\ \hline
Multi-Loss ResNet50                  & 4           & 7.087  & 6.870 & 5.621 & 6.526  \\
                         & 2           & 6.470  & 6.559 & 5.436 & \bf{6.155}  \\
                         & 1           & 6.920  & 6.637 & 5.674 & 6.410  \\
                         & 0.1         & 10.270 & 6.867 & 5.420 & 7.519  \\
                         & 0.01        & 11.410 & 6.847 & 5.836 & 8.031  \\
                         & 0           & 11.628 & 7.119 & 5.966 & 8.238  \\ \hline
Multi-Loss AlexNet                  & 1           & 27.650 & 8.543 & 8.954 & 15.049 \\
                         & 0.1         & 30.110 & 9.548 & 9.273 & 16.310 \\
                         & 0.01        & 25.090 & 8.442 & 8.287 & 13.940 \\
                         & 0           & 24.469 & 8.350 & 8.353 & 13.724
\end{tabular}
}
\caption{Ablation analysis: MAE across different models and regression loss weights on the AFLW2000 dataset.}
\label{AFLW2000_ablation}
\end{table}

\subsection{Low-Resolution AFLW2000 Study}\label{sec4-G}
We study the effects of downsampling all images from the AFLW2000 dataset and testing landmark-to-pose methods on these datasets. We compare these results to our method using different data augmentation strategies.
We test the pose retrieved from the state-of-the-art landmark detection network FAN and also from Dlib. We test all methods on five different scales of downsampling x1, x5, x10 and x15. In general images are around 20-30 pixels wide and high when downsampled x15. We then upsample these images and run them through the detectors and deep networks. We use nearest neighbor interpolation for downsampling and upsampling.

For our method we present a multi-loss ResNet50 with regression coefficient of $1$ trained on normal resolution images. We also train three identical networks: for the first one we augment the dataset by randomly downsampling and upsampling the input image by x10, for next one we randomly downsample and upsample an image by an integer ranging from 1 to 10 and for the last one we randomly downsample and upsample an image by one of the following integers 1, 6, 11, 16, 21.

\begin{figure}[t]
\begin{center}
   \includegraphics[width=1\linewidth]{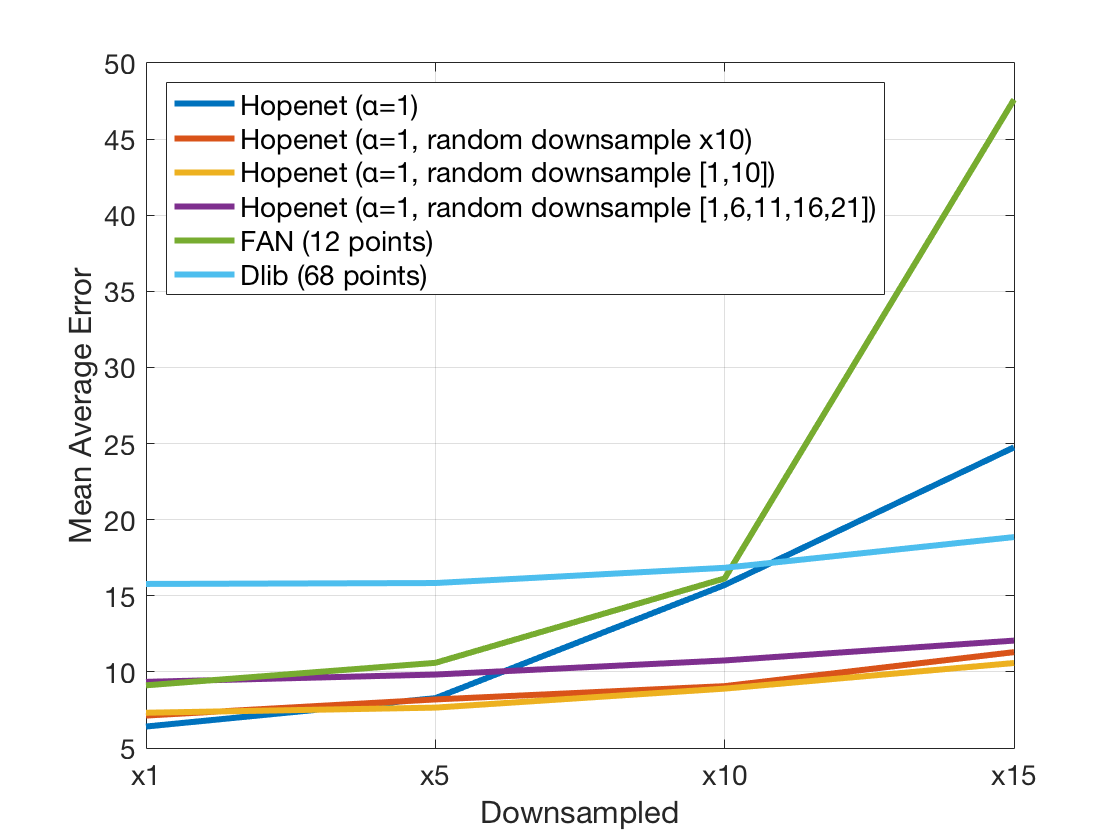}
\end{center}
   \caption{Mean average error for different methods on the downsampled AFLW2000 dataset in order to determine robustness of methods to low-resolution images.}
   \label{low_resolution}
\end{figure}

We observe that from the get-go our methods show better performance than pose from the Dlib landmarks, yet pose from the FAN landmarks is acceptable. Pose from the FAN landmarks degrades as the resolution gets very low which is natural since landmarks are very hard to estimate at these resolutions especially for methods that rely heavily on appearance. Pose from the network without augmentation deteriorates strongly yet the networks with augmentation show much more robustness and perform decently at very low resolutions. Results are presented in Figure \ref{low_resolution}. This is exciting news for long-distance and low-resolution head pose estimation.

\addtolength{\textheight}{-3cm}   

\section{CONCLUSIONS AND FUTURE WORK}\label{sec5}
In this work we show that a multi-loss deep network can directly, accurately and robustly predict head rotation from image intensities. We show that such a network outperforms landmark-to-pose methods using state-of-the-art landmark detection methods. Landmark-to-pose methods are studied in this work to show their dependence on extraneous factors such as head model and landmark detection accuracy.

We also show that our proposed method generalizes across datasets and that it outperforms networks that regress head pose as a sub-goal in detecting landmarks. We show that landmark-to-pose is fragile in cases of very low resolution and that, if the training data is appropriately augmented, our method shows robustness to these situations.

Synthetic data generation for extreme poses seems to be a way to improve performance for the proposed method as are studies into more intricate network architectures that might take into account full body pose for example.


{\small
\bibliographystyle{ieee}
\bibliography{refs}
}

\end{document}